\documentclass{article}

\usepackage{multirow}
\usepackage{tabu}

\usepackage{arxiv}

\usepackage[utf8]{inputenc} 
\usepackage[T1]{fontenc}    
\usepackage{hyperref}       
\usepackage{url}            
\usepackage{booktabs}       
\usepackage{amsfonts}       
\usepackage{nicefrac}       
\usepackage{microtype}      
\usepackage{lipsum}		
\usepackage{graphicx}
\usepackage{natbib}
\usepackage{doi}

\title{Adapting PromptORE for Modern History: Information Extraction from Hispanic Monarchy Documents of the XVI\textsuperscript{th} Century}

\author{Héctor L\'opez Hidalgo \\
  IRIT, UMR5505 CNRS  \\
  Toulouse, France \\
  \texttt{data\_analitics\_HLH@protonmail.com} \\\And
   Michel Boeglin  \\
  IRIEC -- Univ. Paul-Valéry \\
  Montpellier 3, France \\\And
  David Kahn  \\
  INU Jean-Francois Champollion\\
  FRAMESPA, UMR5136 CNRS  \\
  Albi, France  \\
    \texttt{david.kahn@univ-jfc.fr} \\\AND
  Josiane Mothe \\
  Univ. de Toulouse, IRIT, UMR5505 CNRS \\
  Toulouse, France \\
  \texttt{Josiane.Mothe@irit.fr} \\\And
   Diego Ortiz \\
  IRIT, UMR5505 CNRS  \\
  Toulouse, France \\\And
   David Panzoli  \\
  IRIT, UMR5505 CNRS \\
  INU Jean-Francois Champollion \\
  Albi, France\\ \\
    }

\renewcommand{\shorttitle}{\textit{Adapting PromptORE for Modern History} }

\hypersetup{
pdftitle={Adapting PromptORE for Modern History: Information Extraction from Hispanic Monarchy Documents of the XVI\textsuperscript{th} Century},
pdfsubject={Relation extraction, prompt engineering, Spanish historical documents },
pdfauthor={Héctor L\'opez Hidalgo, Michel Boeglin,  David Kahn, Josiane Mothe, Diego Ortiz, David Panzoli },
pdfkeywords={Information systems, Information extraction, Large language models, Prompt engineering, Historical documents },
}

\begin{document}
\maketitle

\begin{abstract}
	Semantic relations among entities are a widely accepted method for relation extraction. PromptORE (Prompt-based Open Relation Extraction) was designed to improve relation extraction with Large Language Models on generalistic documents. However, it is less effective when applied to historical documents, in languages other than English. In this study, we introduce an adaptation of PromptORE  to extract relations from specialized documents, namely digital transcripts of trials from the Spanish Inquisition. Our approach involves fine-tuning transformer models with their pretraining objective on the data they will perform inference. We refer to this process as “biasing”.
Our Biased PromptORE addresses complex entity placements and genderism that occur in Spanish texts. We solve these issues by prompt engineering. We evaluate our method using Encoder-like models, corroborating our findings with experts' assessments. Additionally, we evaluate the performance using a binomial classification benchmark. Our results show a substantial improvement in accuracy --~up to a 50\% improvement with our Biased PromptORE models in comparison to the baseline models using standard PromptORE.
\end{abstract}


\section{Introduction}
Semantic relations between entities such as places, people, and concepts are a popular way to represent knowledge. Relation extraction is the task in natural language processing (NLP) in which relationships between named entities are extracted from textual data. In automatic methods, this process can be implemented as a pipeline where entities are first extracted and then the relationships among them~\cite{pawar2017relation}. Alternatively, it can be implemented as a joint extraction process~\cite{kumar2017survey}. Although the second approach is more challenging, it is also expected to be more effective because dependencies can be better considered~\cite{wang2022deep}. Recent relation extraction methods rely on Large Language Models (LLMs)~\cite{li2020survey}. They have been in use for several years, achieving state-of-the-art performances in most NLP tasks~\cite{rothman2022transformers}. Most of the methods focus on contemporary texts~\cite{xiaoyan2023comprehensive}. These techniques face problems when handling documents of expert's fields, which present unique challenges related to language and syntax. 

Indeed, documents of expert's fields often contain specific language, terms, or idioms that may not be present in  modern LLMs or dictionaries. Modern NLP tools may face difficulties in extracting entities from this type of documents. These difficulties can be observed on legal~\cite{chalkidis2020legal} and medical~\cite{lee2020biobert} domains. As example, medical field has their specific named entity recognizers such as BioBERT~\cite{lee2020biobert}. But documents from other knowledge fields, such as History, may come from a variety of different periods and sources. Even a specific NER may not be generic enough to cover them all. NewsEye project for example~\cite{oberbichler2022integrated} emphasizes the extraction in XIX century newspapers.

Documents of expert's fields may also use a very specific syntax. For example, some historical texts use a punctuation system that was designed for public readers, where comas are intended to help the reader to know where to stop or breath. Some relations may therefore be difficult to extract. This is the case of isolated entities that are generally not considered in the frame of relation extraction techniques, but often appear in documents of expert's fields.

In this paper, we focus on historical documents which combine the properties listed above. The paragraph below is extracted from ``El proceso de Pedro de Cazalla''~\cite{cazalla}, an inquisitorial trial from the 16th century, written in old Castilian and transcribed in the indirect style (customary at the time e.g.: ‘el reo dixo que estaua’ ‘the accused said he was’), with the lexical and syntactic characteristics typical of the Spanish of the time. It illustrates the challenges of the syntax of historical texts. 
\textit{``Y que después de esto, que podría haber un año que el dicho \underline{Padilla} fue a \underline{Pedrosa} a conocer al dicho \underline{Pedro de Cazalla}, y que una tarde el dicho \underline{Pedro de Cazalla} le llevó al dicho \underline{Padilla} a casa de esta confesante, y esta confesante llamó a \underline{Catalina Román}, para que también le conociese, y que allí no se hablaron en ninguna cosa de estas opiniones, por estar su madre de esta confesante delante; y que luego el mismo día, porque él se quiso despedir e irse, esta confesante y la dicha \underline{Catalina Román} se salieron a las eras; tornó a decir que los que se salieron a las eras eran los dichos \underline{Pedro de Cazalla} y \underline{Padilla}, y que al tiempo que \underline{Catalina Román} salieron con ellos, hasta la puerta de la dicha casa, y allí se despidieron con regocijo, dándose a entender, los unos a los otros, que estaban en esto del purgatorio; y que después de esto otra vez fue el dicho \underline{Padilla} a \underline{Pedrosa}, y fue a visitar al dicho \underline{Pedro de Cazalla}, el cual, delante de esta confesante y de la dicha \underline{Catalina Román}, riñó muy de veras al dicho \underline{Padilla}, diciéndole que hablaba muy rotamente de estas cosas con todas las gentes.''}

Since the first grammar of Castilian was written by Nebrija  —published in 1492— until the creation in 1713 of the Real Academia Española (RAE) at the time of the Spanish Enlightenment, Spanish language has experienced important syntactic, lexical,  and prosodic changes, with a system of phoneme accentuation and a standardised system of punctuation. The RAE reformed the orthographic rules of Spanish between 1726 and 1815, producing the majority of modern rules~\cite{petrucci1999alfabetismo}. These systems had a set of rules different from our modern standards. The disposition of dots, commas, and semicolons was merely designed for reading aloud, although nowadays their usage is meant for different purposes. Punctuation in contemporary texts rather serves rhythm, flow and sometimes disambiguation of meaning. 

Nowadays texts favour short sentences, whereas the unique sentence above contains over fifteen entities (underlined in the text), nested within each other. All of this makes relation extraction very challenging, even for the most advanced techniques.  Liu et al.~\cite{liu2023pre} stated that relation extraction methods are typically designed for triplets consisting of two entities and their relationship within one sentence, usually short.

In this paper, we present a step-by-step solution to deal with these complex documents. First we present in section~\ref{sec:Related Work} the state-of-the-art methods upon which we base our original approach.
Section~\ref{sec:methodology} introduces our research methodology and our hypothesis of work.
Then, a technical description of the technique is described in section~\ref{sec:technique}.
Finally, we present and discuss our findings in section~\ref{sec:Results} and give our conclusions in section~\ref{sec:Conclusion}.

\begin{figure}
\centering
\includegraphics[width=0.48\textwidth]{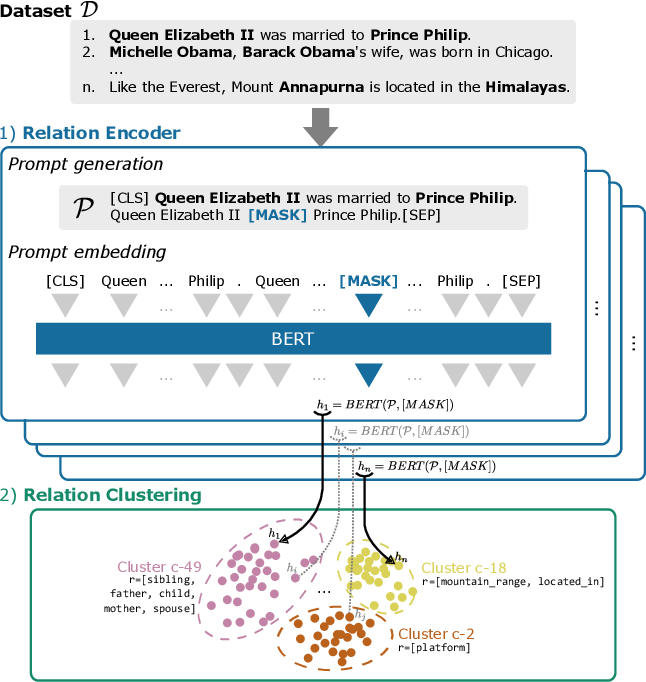}
\caption{Original PromptORE technique overview, extracted from~\cite{genest2022promptore}. Based on a dataset $\mathcal{D}$ where each phrase contains exactly two entities, the authors encode the relationships with a Mask Language Modelling objective. Then, the [MASK] token needs to be guessed by the model, reducing the cross-entropy loss function, and learning the semantics of the language in the process.}
\label{fig:prompt}
\end{figure}

\section{Related Work}\label{sec:Related Work}
Since 2019, LLMs such as BERT, RoBERTa or GPT have been used as state-of-the-art methods for relation extraction in English~\cite{han2020novel,gharagozlou2023semantic}. One trending technique involves the use of prompts, which are instructions provided in  natural language to an LLM for the extraction of entities and relations. This is known as prompt engineering~\cite{liu2023pre}. Prompt engineering takes advantage of the inherent semantic knowledge within these LLMs. It demonstrates that general distributions of languages captured by an LLMs, when combined with an effective prompt, can successfully extract meaningful relationships between entities. 

Within the variety of prompt engineering techniques to perform relation extraction~\cite{liu2023pre}, most of them employ the Masked Language Modeling (MLM) objective, which is characteristic of BERT-like models, as in Figure~\ref{fig:prompt}. MLM consists in corrupting some tokens with the special token [MASK]. The model needs to reduce the cross-entropy loss function guessing the corrupted token. This is quite effective in relation extraction, as it transforms the task into a prompt engineering problem. Some authors include other special continuous tokens to influence and improve their prompt~\cite{son2022grasp}. Others use token positions and tagging techniques~\cite{chen2022knowprompt}. But, to the best of our knowledge, PromptORE~\cite{genest2022promptore} is the only one that has the advantage of not requiring model fine-tuning to perform relation extraction.

PromptORE technique is illustrated in Figure~\ref{fig:prompt}. Based on a dataset $\mathcal{D}$ where each sentence contains exactly two entities, the relationships are encoded with a prompt structure $\mathcal{P}$. Then, the [MASK] token hidden state, the last computation of the BERT encoder ($\mathcal{P}$, [MASK]), is compared with all the other [MASK] hidden states across all sentences using cosine similarity to cluster the relationships. The technique takes advantages of the prompting design, which is key in modern relation extraction performed with transformer models. The authors used BERT MLM pretrained objective without fine-tuning.

Regarding detailed approaches on extraction for historical documents, most studies rely on documents that, chronologically, belong to the XIX\textsuperscript{th} and XX\textsuperscript{st} centuries, such as~\cite{oberbichler2022integrated}. 
New approaches~\cite{gonzalez2023yes} with GPT models remain in this chronology, as it has better and more sources than prior periods. Therefore, there are no articles to the best of our knowledge that deals with the complexity of prior periods, such as Modern Age. 



Despite their great effectiveness, relation extraction techniques often overlook relation and entity extraction challenges such as isolated entities, long-range dependency relationships between entities, and specific semantic nuances that are inherent to texts from before the XIX\textsuperscript{th} century.

Based on the PromptORE approach, and considering the unique challenges presented by historical documents as specialized domain, we think that additional pretraining of LLMs, focused on a particular semantic field, could enhance the effectiveness of prompt engineering. In the next sections, we present an original approach that consists of ``biasing'' LLM models and that we have called the bias-prompt-extract technique.

\section{Methodology}\label{sec:methodology}

Historical documents in Spanish open new challenges that are not yet solved. We developed  a new approach based on  prompt engineering for relation extraction in this context. 

We hypothesize that biasing the models towards the  target knowledge domain for relation extraction will improve the results of prompts to deal both with difficult entity positions inside texts and gender considerations of Spanish language. We also hypothesize that isolated entities can be extracted  by creating a specific prompt. We name this idea ``anaphoric prompting''. 

\subsection{Overview}

Using promptORE on English and general domain documents demonstrated that encoder-like LLMs do not need to be fined-tuned to extract relations~\cite{genest2022promptore}. Therefore, biasing a pretrained model towards a semantic field appears as a potentially appropriate solution to improve relation extraction on documents from specific fields or languages. 

\subsection{PromptORE}
\label{sub:promptORE}

PromptORE technique is illustrated in Figure~\ref{fig:prompt}. Based on a dataset $\mathcal{D}$ where each sentence contains exactly two entities, the relationships are encoded with a prompt structure $\mathcal{P}$. Then the [MASK] token hidden state, the last computation of the BERT encoder ($\mathcal{P}$, [MASK]), is compared with all the other [MASK] hidden states across all sentences using cosine similarity to cluster the relationships without human intervention. The technique takes advantages of the prompting design, which is key in modern relation extraction performed with transformer models. The authors use BERT masked language modeling pretrained objective without fine-tuning.
Their model was designed for English. 

    

The promptORE  design  turns out to be unpractical for Spanish. Indeed, the initial prompt cannot deal with gendered languages since the system was designed for English. When we used it on our preliminary tests on the Spanish historical documents, the outcomes generated by pretrained transformers in Spanish predominantly consisted of punctuation symbols (e.g. the most probable tokens for different cases were ";", "," or "y" (and), "o" (or) when asked for the relation between entities.  Grammatically speaking they are correct, yet uninformative. Indeed, with BETO~\cite{canete2023spanish}, a BERT encoder for Spanish trained by University of Santiago de Chile, the technique mostly fetched punctuation symbols, which are correct inferences for [MASK], but irrelevant with respect to the expected prediction which should be a verb for instance. With Spanish RoBERTa Base~\cite{gutierrez2021maria}, the technique provides slightly more informative inferences. et, propositions are merely as informative as punctuation symbols. With Spanish language, the prompts need to be re-designed.  

The authors of PromptORE evaluated it on FewRel~\cite{han2018fewrel}. This data set consists of triplets of entities and relationships from a general variety of topics with news from several media sources  described by the authors as  specialised content data. Although the data set is specific in that it contains sentences related to specific topical issues, these sentences and their relationships are not specialised  at the level  historical documents are.

\subsection{Hypothesis of work}\label{challenges}
Within the family of LLMs, the primary pretraining objective of an encoder model is to learn the language general representation~\cite{vaswani2017attention}. This involves comprehending its semantics, grammar, and other relevant properties. We selected BERT~\cite{devlin2018bert} and RoBERTa~\cite{liu2019roberta}, as they are the most used LLM encoder models for relation extraction. We chose PromptORE technique as the baseline for relation extraction;  it has the advantage of being versatile and having a simple set-up. 

PromptORE technique considers sentences that contain exactly two entities. It also assumes that the sentence contains the relationship between the two entities. BERT model encodes this with a structured prompt (see Section~\ref{sub:promptORE}). The hidden states of all [MASK] tokens resulting from the final computation of the BERT encoder are compared using cosine measure. This comparison is conducted by an unsupervised algorithm like K-means, which clusters and classifies the relationships automatically, 
considering a predefined number of clusters K. We find the idea of using the embedding computation with K-means useful, as using embeddings for Social Science relation extraction is a common and successful practice~\cite{canete2023spanish}.

Sentences in historical documents do not follow the same structure as in the FewRel dataset used to evaluate PromptORE. To demonstrate the versatility of our approach, we use an ad hoc historical benchmark dataset  (see Section~\ref{sec:technique}). 

To better extract relations in historical documents, we aim to bias BERT and RoBERTa with the MLM learning objective using a corpus of texts centered on a specific chronological period and semantic field. We consider the problem as a task of relation extraction triplets~\cite{xiaoyan2023comprehensive} and as a problem of prompting design in Spanish, considering the following challenges:
\begin{itemize}
    \item The use of gender in the Spanish language poses challenges that do not exist in the original technique designed for  English.
    \item In our corpus, cases are not  limited to two entities per sentence: most contain more, while some include only one entity.
    \item Isolated entities sentences, where the entity has a relationship with the document itself, which is particular to legal documents in the Modern Age (see sub-section~\ref{sec:compo}).
\end{itemize}

\begin{figure*}
    \centering
    \includegraphics[width=\textwidth]{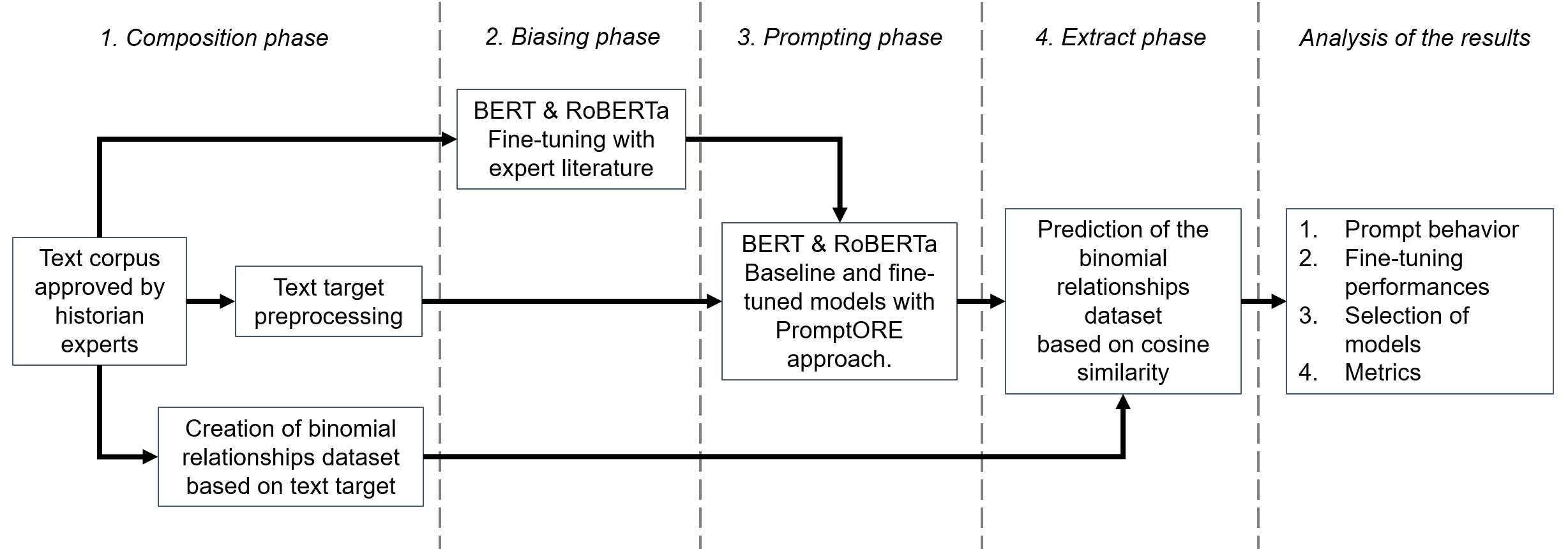}
    \caption{An overview of our bias-prompt-extract technique for complex historical documents. The biasing phase is inserted between the composition phase (selection of documents from the period) and the prompting phase (testing PromptORE baselines, biased BERT and RoBERTa models.}
    \label{fig:overview}
\end{figure*}

In order to address these issues, the Biased PromptORE technique is structured in four phases, illustrated in Figure~\ref{fig:overview}.
The composition phase is where the data are prepared for the next steps. It consists of building-up a binomial dataset for the final benchmark and selecting texts that will be useful for the biasing phase.
During the biasing phase, we apply the MLM objective of BETO and RoBERTa models in Spanish to create a biased model using the data previously complied for this usage at the composition phase.
The prompting phase takes advantage of our newly developed prompt design, which improves the original PromptORE for Spanish documents of the Modern Age.
Finally, the extraction phase consists in comparing the results of the Biased PromptORE with the baseline PromptORE model on the binomial dataset created in the composition phase. 

\section{Biased-PromptORE technique}\label{sec:technique}

We thoroughly describe each phase of the bias-prompt-extract technique introduced  Section~\ref{sec:methodology}.

\subsection{Composition phase}\label{sec:compo}
 
To bias the LLMs using MLM, we created a corpus centered on the same semantic domain (inquisitorial Spanish religious history of the XVI\textsuperscript{th} century) with the help of historians. It consists of 5 long (12,661 words in average) historical documents, ranging from legal records (trials of heresy by the Spanish Inquisition) to religious literature (Catholic doctrine books). We divided the corpus between the target text and expert books. The target text is the one on which we want to extract relations while the expert books are used to fine-tune BERT and RoBERTa models. 

\paragraph{Target text}
The target text is an inquisitorial trial popularly known as ``El Proceso de Pedro de Cazalla''~\cite{cazalla}. The historians chose this document because it is quite rich on named entities and complex relationships, including the issues we mentioned in the methodology section. The document was normalized, by removing word contractions and modernizing terms. Also named entities were extracted by the historian team. We also use this document in the second dataset to show that knowing the text beforehand  improves  relation extraction.

In addition to  specific features of Spanish texts from the Modern Age (stated in section~\ref{challenges}), this type of legal document also includes peculiar sentences where isolated entities have a relation with the document itself. For example, in the following sentence of the Cazalla trial, \emph{``Pasó ante mí; Sebastián de Landeta, Notario.''} (He/she testified before me; Sebasti\'an de Landeta, Notary.), the notary named Sebasti\'an de Landeta confirms that the testimony was indeed given by the witness. 

\begin{figure*}
    \centering
    \includegraphics[width=\textwidth]{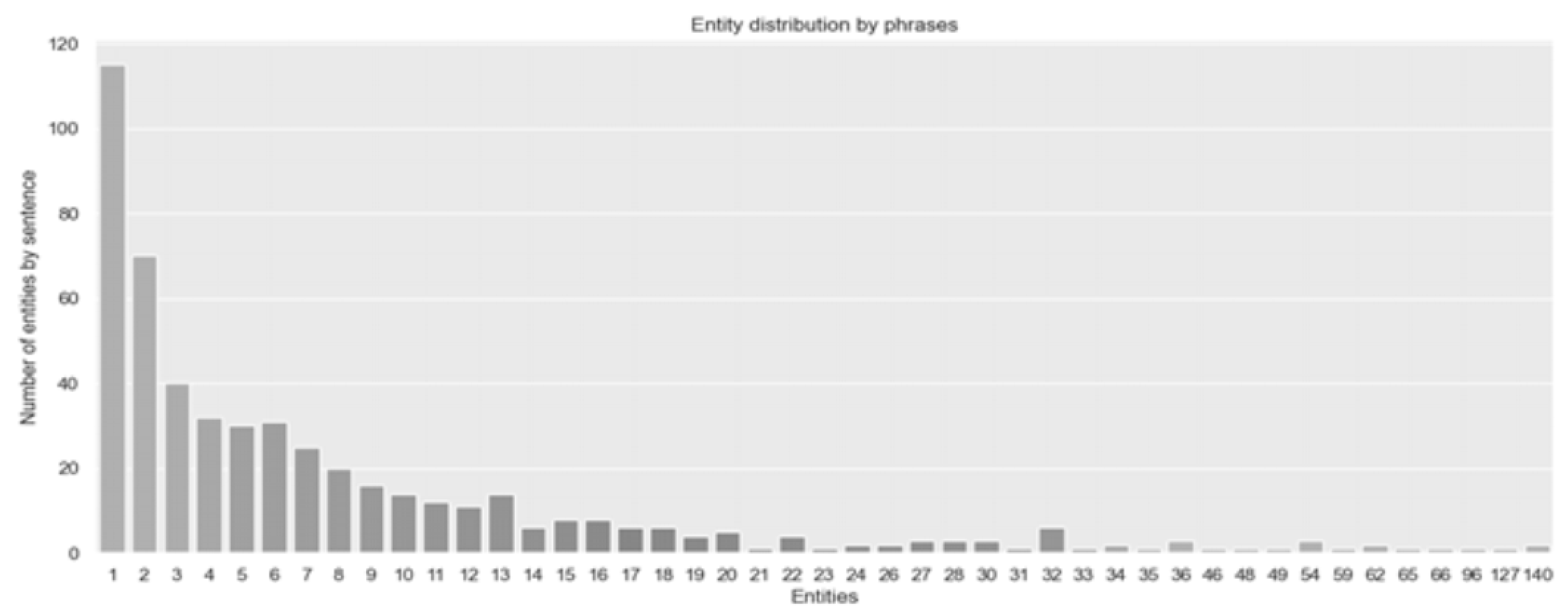}
    \caption{\label{fig:distribution}Number of entities per sentence:  splitting the sentences by  the punctuation marks revealed the complexity of parsing texts: a significant number of sentences include too many entities for usual tools and  humans to be able to draw any reliable information from them.}
\end{figure*}

Figure~\ref{fig:distribution} reports the breakdown of the number of entities per sentence. We can observe that many sentences  have more than five entities; even, two sentences have more than 130 entities each. These outliers make it almost impossible to parse and divide the whole text by sentence like in modern texts. We also observed that 120 sentences contain a single entity that has a relationship with the document itself in the trial.
We found that 75 percent of the sentences have less 4 entities. We focused on these sentences with 1, 2 or 3 entities for the target text.

\paragraph{Expert books}
The expert books consist in a collection of different manuals and expert articles to bias BERT and RoBERTa models:

- The bible: we used a popular reformist bible of the XVI\textsuperscript{th} century called ``La Biblia del Oso''. Historians recommended this tome for MLM fine-tune towards the semantic general fields that are going to appear in our target text. The tokens from this document follow almost a normal distribution and it is the longest, therefore, quite appropriated for the biasing phase.

- Ecclesiastic and inquisitorial concept dictionaries. These dictionaries contain expert definitions and semantic related fields that will be used to bias the models.

- The trial of Cazalla\footnote{Note that the trial of Cazalla is used both as target text and as part of the expert books. We hypothesize the structure and grammar of this text will help in the extraction task.} and the Catechism of Pineda: there are XVI\textsuperscript{th} century legal and religious texts, respectively. Both share the same vocabulary and semantic fields. 

We created a binomial dataset based on the target text for the extract phase, and we pre-processed the expert books for the biasing phase.



\begin{table}[!htb]
\caption{Distribution of words by sentence in the Expert Books data set. }
\noindent
\begin{tabular}{lllll}
\hline
\textbf{Title} & \textbf{Mean} & \textbf{Std} & \textbf{Median} & \textbf{Total} \\
\hline 
Bible book & 23.97 & 20.04 & 22 & 31,095 \\
Ecclesiastical Dictionary & 19.83 & 14.29 & 16.00 & 17,543 \\
Catechism of Pineda & 20.04 & 14.06 & 16 & 892 \\
Process of Cazalla & 77.13 & 105 & 48 & 1115 \\
\hline
\end{tabular}
\label{tab:distribution}
\end{table}

\subsection{Biasing phase}
We used RoBERTa and BETO encoder models that are already pretrained for Spanish. We selected three models : BETO~\cite{canete2023spanish}, RoBERTa base and RoBERTa large~\cite{gutierrez2021maria}. BETO is a fully trained BERT model in Spanish developed by the University of Computation of Santiago de Chile. RoBERTa base and large in Spanish were trained within the MarIA project~\cite{gutierrez2021maria}.
We biased the models with the expert books data set. We present the training settings, such as the learning rate of 5e-5 for BETO and RoBERTa base and 5e-6 for RoBERTa Large. All of them were trained for 5 epochs. the losses, for BETO were 0.0484, for RoBERTa base 0.0627 and for RoBERTa Large 0.9899. In this, we follow what has been done in earlier studies~\cite{devlin2018bert,liu2019roberta,canete2023spanish,gutierrez2021maria}.

\subsection{Prompting phase}

The original PromptORE technique aims to design the simplest prompt possible.
For a given dataset $\mathcal{D}$, composed of $S$ sentences as in Equation 1, given sentence $s$ as in Equation 2, a sentence with 2 entities ($e_1$, $e_2$) and one relation ($r$), the design, represented in Equation 3, is the simplest possible prompt. We require BERT and RoBERTa encoders to predict a grammatically correct word, which in English would be a verb~\cite{genest2022promptore}.


\begin{equation}
\mathcal{D}= S = [s_1,s_2,…,s_n]
\end{equation}
\begin{equation}
s=(e_1,e_2,r)
\end{equation}
\begin{equation}
\mathcal{P}=(s,e_1,[MASK],e_2)
\end{equation}

There is no specific prompt designed for Spanish using PromptORE technique~\cite{geron2020aprende}. We created five prompts (Equ. 4--8). 
We compared the original PromptORE prompts, which do not produce fixed-gendered words, with prompts that specifically include fixed-gendered terms.
%
Equ. 4--6 show the original PromptORE  with  non-fixed gendered prompts for Spanish whereas Equ. 7--8 show the fixed gendered prompts.

\begin{equation}
\mathcal{P}_0=(s,e_1,[MASK],e_2,[SEP])
\end{equation}
\begin{equation}
\mathcal{P}_1=(s,\textrm{``la relación entre''},e_1,\textrm{``y''},e_2, \\ \textrm{``es una relación de''},[MASK],[SEP])
\end{equation}
\begin{equation}
\mathcal{P}_2=(s,\textrm{``la relación entre''},e_1,\textrm{``y''},e_2, \\ \textrm{``es''},[MASK],[SEP])
\end{equation}
\begin{equation}
\mathcal{P}_3=(s,\textrm{``la relación entre''},e_1,\textrm{``y''},e_2, \\ \textrm{``es la''},[MASK],[SEP])
\end{equation}
\begin{equation}
\mathcal{P}_4=(s,\textrm{``la relación entre''},e_1,\textrm{``y''},e_2, \\ \textrm{``es el''},[MASK],[SEP])
\end{equation}

There are several sentences on which one entity was isolated, but based on a manual analysis by historians, there are relationships between entities and the text itself (for example: "Paso ante mi, Sebastián Landeta." \textit{Step before me, Sebastian Landeta.}). We need a specific type of prompting design to extract these relationships. We designed an anaphoric prompt, which we use to evaluate the capabilities of BERT and RoBERTa with hidden relationships that are obvious by the context for a human analysis (See Equ. 9). We created a “phantom” entity, which refers contextually to the sentence itself. 

\vspace{-4pt}
\begin{equation}
\mathcal{P}_{Anaphoric}=(s,\textrm{``la relación entre''} ,e_1, \textrm{``y la frase} \\ \textrm{anterior es una relación de''} ,[MASK],[SEP])
\end{equation}
\vspace{-4pt}

\begin{table*}
\caption{Model outputs presented to the historians and based on our target text. In bold the models that were selected for the final binomial classification task.}
\noindent
\begin{tabular}{lllll}
\hline
\textbf{Model} & \textbf{Output} \\
\hline 
\multirow{2}{*}{Bert (BETO)} & Baseline	& [``cosas'', ``palabras'', ``ellas'', 
``ella'', ``dios'']\\
& \textbf{Biased}	& [``derecho'', ``muerte'', ``paz'', ``ellas'', ``importancia'']\\
\tabucline{2-}
\multirow{2}{*}{RoBERTa Base} & Baseline	& [``palabras'', ``hechos'', ``errores'', ``dos'', ``nombres'']\\
& \textbf{Biased}	& [``fiscal'', ``familiar'', ``privado'', ``particular'', ``tributario'']\\
\tabucline{2-}
\multirow{2}{*}{RoBERTa Large} &  Baseline	& [``muy'', ,``privado'', ``público'', ``personal'', jurídico'']\\
& Biased	& [``siguiente'', ``misma'', ``de'', ``presente'', ``dicha:'']\\
\hline
\end{tabular}
\label{tab:outputs} 
\end{table*}

Finally, to deal with three entities within a sentence, we choose a combinatory strategy without repetition. This allows us to implement data augmentation for the corpus, multiplying by 3 the number of sentences with 3 entities.

\subsection{Relation extraction phase}
PromptORE technique, as we explained earlier, takes advantage of the MLM objective of BETO and RoBERTa base~\cite{devlin2018bert,liu2019roberta} to predict the most semantically plausible word for the [MASK] token. After, the final hidden state of that mask token, which is the last computation the models perform over the embedding of the token, it is taken and compared with cosine similarity by a non-supervised classifier as K-means clustering~\cite{genest2022promptore}. 

Before clustering the hidden states, in our experiments, the historians checked the outputs of the models to select the most semantically accurate word. We took the ten most probable words predicted by the models and asked the historians to select the 2 models (one for BETO, the other for RoBERTa) to be tested with the binomial classification benchmark (See Table~\ref{tab:outputs} ). The baseline model had the natural tendency to generalize the context due to their pretrain objective. On the other hand, the biased models were more specific in the selection of words. Finally, the large versions of RoBERTa performed poorly. One of the reason could be  the model has too many parameters considering the relatively low amount of  data used to bias them. Also, the further we biased the model, the closest it came to the semantic field.

For the binomial classification task, we compared the baseline PromptORE and our Biased PromptORE combining normal prompt types with anaphoric prompts.
We test every $\mathcal{P}_n$ with both models. $\mathcal{P}_1$ to $\mathcal{P}_4$ work for 2-3 entities and $\mathcal{P}_{Anaphoric}$  captures 1 entity sentences. A complete test should include any $\mathcal{P}_n$  plus $\mathcal{P}_{Anaphoric}$.  Then, we clustered the selected model outputs with  K-means clustering. We set K=2 to test our approach on the binomial dataset.

\section{Results}\label{sec:Results}

\subsection{Prompting results}
We test every prompt $\mathcal{P}_n$ with both models. $\mathcal{P}_1$ to $\mathcal{P}_4$ work for 2-3 entities and $\mathcal{P}_{Anaphoric}$  captures 1 entity sentences. A complete test should include any $\mathcal{P}_n$  plus $\mathcal{P}_{Anaphoric}$.  Then, we clustered the selected model outputs with  K-means clustering. We set K=2 to test our approach on the binomial dataset. $\mathcal{P}_1$ was the most successful of the prompts (See Table~\ref{tab:outputs}). It handles genders yielding a rich, accurate semantic selection of words. BETO gives also the most consistent results with $\mathcal{P}_1$. $\mathcal{P}_2$  is quite successful when focusing on the semantic field. Nevertheless, $\mathcal{P}_2$   output has a greater degree of randomness than $\mathcal{P}_1$.
$\mathcal{P}_3$ and $\mathcal{P}_4$  are successful regarding genderism within the semantic field. Some words produced by the model are better than $\mathcal{P}_1$ and $\mathcal{P}_2$ because they specify a feminine or masculine relationship. However, generating only masculine or feminine results would be a limitation of our prompting design.  
The Anaphoric prompt is appropriate for this type of legal text; it correctly identifies the relationships; this result is consistent in the four tests performed. The “phantom entity”  allows the biased models to catch the semantic relationship and cluster it accurately. 

The prompts we designed achieve relation extraction in Spanish, specifically for ancient texts. $\mathcal{P}_1$ is a general and stable option for the biased models, $\mathcal{P}_3$ and $\mathcal{P}_4$ is useful in contexts where the relationship between entities can be define as grammatically feminine or masculine in Spanish.

\begin{table*}
    \caption{Results with the binomial dataset from our Target Text. Best results are in bold font. }
    \label{tab:results}
    \begin{tabular}{lllll}
    \hline
    \textbf{Model} &  \textbf{Accuracy} & \textbf{Precision} & \textbf{Recall} & \textbf{F1 Score} \tabularnewline
    \hline 
BERT (BETO) baseline	& 0.1569 & 0.0641 & 0.0292 & 0.040\tabularnewline
BERT (BETO) biased	& \textbf{0.6578} & \textbf{0.6456} & \textbf{0.5795} & \textbf{0.7717}\tabularnewline
RoBERTa Base baseline	& 0.2945 & 0.3259 & 0.3551 & 0.3399\tabularnewline
RoBERTa Base biased	& 0.4585 & 0.4688 & 0.4413 & 0.4547\tabularnewline
\hline
\end{tabular}
\end{table*}

\subsection{Relation Extraction results}
Table~\ref{tab:results}  reports the comparison between the biased PromptORE models we developed and the PromptORE considered as baselines. As we can see, baseline BETO and both, RoBERTa base and large baselines, performed poorly. The outputs predicted for the [MASK] token in the classification were too general, giving as results low accuracy, precision, recall and F-1 scores.

Biased PromptORE models are much more effective, whatever the metrics we consider. Biased BETO consistently outperforms the other models. Also, RoBERTa Base is improved when biased, whatever the measure. Biased PromptORE RoBERTa base improves by an average of 20\% all its scores compared to the baseline. The best model, Biased PromptORE BETO, improves accuracy up to 50\%, resulting in an overall F1 score of 77\%. This model was also selected by the historians as the model to use for relation extraction because of its performance and accuracy. This model, combined with $\mathcal{P}_1$  and $\mathcal{P}_anaphoric$ prompts  made much more easier relation extraction within the target text. 
We found that using combinatory rules with no repletion for the sentences that contain 3 entities also works quite well, as the model can accurately predict different types of relationships among entities within the same sentence. Nevertheless this approach will yield incorrect results for more than 3 entities within the same sentence. 
Having more than 3 entities may yield uninformative or non accurate results because the further apart the entities are, the less likely they have a meaningful relationship within the text.


For biased RoBERTa Large, its poor performance may be due to  the trade-off between the relatively small amount of data used and the size of the model itself, which may be too large to properly being biased as mentioned in~\cite{geron2020aprende}.

When analysing the  confusion matrices for both PromptORE baselines and biased PromptORE models, the results show that the results were indeed much imrpoved for both models. This also shows the possibility of biasing, and fine-tune in a downstream task as quite preferable for complex historical documents. For the baselines, many relationships are misclassified. The general knowledge of Spanish of these models allows them to be generalist only and appear to be inappropriate for the task. But, when we bias them, we obtain better results for both BETO and RoBERTa outperforming the baselines by far, and also offering a rich and accurate prediction for the [MASK] token. 

There is still room for improvement. BETO is too sure of everything being a legal relationship, preferring them as type. The reason behind this is that some of the sentences are too specific on their selection of semantic context that, for a non-expert,  would look like a legal relationship. But, knowing that these models have no further fine-tune for relation extraction gives us other possibilities, as further fine-tune the models to process the dataset.

\section{Conclusions}\label{sec:Conclusion}
As we presented throughout this paper, our biased PromptORE modification improves the relation extraction for documents from expert fields, specifically historical ones, and our design of prompt outperforms the simple from \mbox{PromptORE} for Spanish language. Our goal now is to achieve better performance in a multiclass dataset benchmark, also composed by historical documents.
The problem of dealing with relation extraction beyond the triplet method still remains and important field of research that other authors tried to solve.


\begin{thebibliography}{25}
\providecommand{\natexlab}[1]{#1}

\bibitem[{caz()}]{cazalla}

\newblock Proceso de fe de Pedro de Cazalla. Archivo Histórico Nacional (España), Sección inquisición, legajo 1864, expediente 2.

\bibitem[{Ca{\~n}ete et~al.(2023)Ca{\~n}ete, Chaperon, Fuentes, Ho, Kang, and P{\'e}rez}]{canete2023spanish}
Jos{\'e} Ca{\~n}ete, Gabriel Chaperon, Rodrigo Fuentes, Jou-Hui Ho, Hojin Kang, and Jorge P{\'e}rez. 2023.
\newblock Spanish pre-trained bert model and evaluation data.
\newblock \emph{arXiv preprint arXiv:2308.02976}.

\bibitem[{Chalkidis et~al.(2020)Chalkidis, Fergadiotis, Malakasiotis, Aletras, and Androutsopoulos}]{chalkidis2020legal}
Ilias Chalkidis, Manos Fergadiotis, Prodromos Malakasiotis, Nikolaos Aletras, and Ion Androutsopoulos. 2020.
\newblock Legal-bert: The muppets straight out of law school.
\newblock \emph{arXiv preprint arXiv:2010.02559}.

\bibitem[{Chen et~al.(2022)Chen, Zhang, Xie, Deng, Yao, Tan, Huang, Si, and Chen}]{chen2022knowprompt}
Xiang Chen, Ningyu Zhang, Xin Xie, Shumin Deng, Yunzhi Yao, Chuanqi Tan, Fei Huang, Luo Si, and Huajun Chen. 2022.
\newblock Knowprompt: Knowledge-aware prompt-tuning with synergistic optimization for relation extraction.
\newblock In \emph{Proceedings of the ACM Web conference 2022}, pages 2778--2788.

\bibitem[{Devlin et~al.(2018)Devlin, Chang, Lee, and Toutanova}]{devlin2018bert}
Jacob Devlin, Ming-Wei Chang, Kenton Lee, and Kristina Toutanova. 2018.
\newblock Bert: Pre-training of deep bidirectional transformers for language understanding.
\newblock \emph{arXiv preprint arXiv:1810.04805}.

\bibitem[{Genest et~al.(2022)Genest, Portier, Egyed-Zsigmond, and Goix}]{genest2022promptore}
Pierre-Yves Genest, Pierre-Edouard Portier, El{\"o}d Egyed-Zsigmond, and Laurent-Walter Goix. 2022.
\newblock Promptore-a novel approach towards fully unsupervised relation extraction.
\newblock In \emph{Proceedings of the 31st ACM International Conference on Information \& Knowledge Management}, pages 561--571.

\bibitem[{G{\'e}ron(2020)}]{geron2020aprende}
Aur{\'e}lien G{\'e}ron. 2020.
\newblock Aprende machine learning con scikit-learn, keras y tensorflow.
\newblock \emph{Espa{\~n}a: Anaya}.

\bibitem[{Gharagozlou et~al.(2023)Gharagozlou, Mohammadzadeh, Bastanfard, and Ghidary}]{gharagozlou2023semantic}
Hamid Gharagozlou, Javad Mohammadzadeh, Azam Bastanfard, and Saeed~Shiry Ghidary. 2023.
\newblock Semantic relation extraction: A review of approaches, datasets, and evaluation methods with looking at the methods and datasets in the persian language.
\newblock \emph{ACM Transactions on Asian and Low-Resource Language Information Processing}, 22(7):1--29.

\bibitem[{Gonz{\'a}lez-Gallardo et~al.(2023)Gonz{\'a}lez-Gallardo, Boros, Girdhar, Hamdi, Moreno, and Doucet}]{gonzalez2023yes}
Carlos-Emiliano Gonz{\'a}lez-Gallardo, Emanuela Boros, Nancy Girdhar, Ahmed Hamdi, Jose~G Moreno, and Antoine Doucet. 2023.
\newblock Yes but.. can chatgpt identify entities in historical documents?
\newblock \emph{arXiv preprint arXiv:2303.17322}.

\bibitem[{Guti{\'e}rrez-Fandi{\~n}o et~al.(2021)Guti{\'e}rrez-Fandi{\~n}o, Armengol-Estap{\'e}, P{\`a}mies, Llop-Palao, Silveira-Ocampo, Carrino, Gonzalez-Agirre, Armentano-Oller, Rodriguez-Penagos, and Villegas}]{gutierrez2021maria}
Asier Guti{\'e}rrez-Fandi{\~n}o, Jordi Armengol-Estap{\'e}, Marc P{\`a}mies, Joan Llop-Palao, Joaquin Silveira-Ocampo, Casimiro~Pio Carrino, Aitor Gonzalez-Agirre, Carme Armentano-Oller, Carlos Rodriguez-Penagos, and Marta Villegas. 2021.
\newblock Maria: Spanish language models.
\newblock \emph{arXiv preprint arXiv:2107.07253}.

\bibitem[{Han and Wang(2020)}]{han2020novel}
Xiaoyu Han and Lei Wang. 2020.
\newblock A novel document-level relation extraction method based on bert and entity information.
\newblock \emph{Ieee Access}, 8:96912--96919.

\bibitem[{Han et~al.(2018)Han, Zhu, Yu, Wang, Yao, Liu, and Sun}]{han2018fewrel}
Xu~Han, Hao Zhu, Pengfei Yu, Ziyun Wang, Yuan Yao, Zhiyuan Liu, and Maosong Sun. 2018.
\newblock Fewrel: A large-scale supervised few-shot relation classification dataset with state-of-the-art evaluation.
\newblock \emph{arXiv preprint arXiv:1810.10147}.

\bibitem[{Kumar(2017)}]{kumar2017survey}
Shantanu Kumar. 2017.
\newblock A survey of deep learning methods for relation extraction.
\newblock \emph{arXiv preprint arXiv:1705.03645}.

\bibitem[{Lee et~al.(2020)Lee, Yoon, Kim, Kim, Kim, So, and Kang}]{lee2020biobert}
Jinhyuk Lee, Wonjin Yoon, Sungdong Kim, Donghyeon Kim, Sunkyu Kim, Chan~Ho So, and Jaewoo Kang. 2020.
\newblock Biobert: a pre-trained biomedical language representation model for biomedical text mining.
\newblock \emph{Bioinformatics}, 36(4):1234--1240.

\bibitem[{Li et~al.(2020)Li, Sun, Han, and Li}]{li2020survey}
Jing Li, Aixin Sun, Jianglei Han, and Chenliang Li. 2020.
\newblock A survey on deep learning for named entity recognition.
\newblock \emph{IEEE Transactions on Knowledge and Data Engineering}, 34(1):50--70.

\bibitem[{Liu et~al.(2023)Liu, Yuan, Fu, Jiang, Hayashi, and Neubig}]{liu2023pre}
Pengfei Liu, Weizhe Yuan, Jinlan Fu, Zhengbao Jiang, Hiroaki Hayashi, and Graham Neubig. 2023.
\newblock Pre-train, prompt, and predict: A systematic survey of prompting methods in natural language processing.
\newblock \emph{ACM Computing Surveys}, 55(9):1--35.

\bibitem[{Liu et~al.(2019)Liu, Ott, Goyal, Du, Joshi, Chen, Levy, Lewis, Zettlemoyer, and Stoyanov}]{liu2019roberta}
Yinhan Liu, Myle Ott, Naman Goyal, Jingfei Du, Mandar Joshi, Danqi Chen, Omer Levy, Mike Lewis, Luke Zettlemoyer, and Veselin Stoyanov. 2019.
\newblock Roberta: A robustly optimized bert pretraining approach.
\newblock \emph{arXiv preprint arXiv:1907.11692}.

\bibitem[{Oberbichler et~al.(2022)Oberbichler, Boro{\c{s}}, Doucet, Marjanen, Pfanzelter, Rautiainen, Toivonen, and Tolonen}]{oberbichler2022integrated}
Sarah Oberbichler, Emanuela Boro{\c{s}}, Antoine Doucet, Jani Marjanen, Eva Pfanzelter, Juha Rautiainen, Hannu Toivonen, and Mikko Tolonen. 2022.
\newblock Integrated interdisciplinary workflows for research on historical newspapers: Perspectives from humanities scholars, computer scientists, and librarians.
\newblock \emph{Journal of the Association for Information Science and Technology}, 73(2):225--239.

\bibitem[{Pawar et~al.(2017)Pawar, Palshikar, and Bhattacharyya}]{pawar2017relation}
Sachin Pawar, Girish~K Palshikar, and Pushpak Bhattacharyya. 2017.
\newblock Relation extraction: A survey.
\newblock \emph{arXiv preprint arXiv:1712.05191}.

\bibitem[{Petrucci(1999)}]{petrucci1999alfabetismo}
Armando Petrucci. 1999.
\newblock Alfabetismo, escritura, sociedad, barcelona, gedisa.
\newblock \emph{PMCid: PMC1905197}.

\bibitem[{Rothman and Gulli(2022)}]{rothman2022transformers}
Denis Rothman and Antonio Gulli. 2022.
\newblock \emph{Transformers for Natural Language Processing: Build, train, and fine-tune deep neural network architectures for NLP with Python, PyTorch, TensorFlow, BERT, and GPT-3}.
\newblock Packt Publishing Ltd.

\bibitem[{Son et~al.(2022)Son, Kim, Lim, and Lim}]{son2022grasp}
Junyoung Son, Jinsung Kim, Jungwoo Lim, and Heuiseok Lim. 2022.
\newblock Grasp: Guiding model with relational semantics using prompt.
\newblock \emph{arXiv preprint arXiv:2208.12494}.

\bibitem[{Vaswani et~al.(2017)Vaswani, Shazeer, Parmar, Uszkoreit, Jones, Gomez, Kaiser, and Polosukhin}]{vaswani2017attention}
Ashish Vaswani, Noam Shazeer, Niki Parmar, Jakob Uszkoreit, Llion Jones, Aidan~N Gomez, {\L}ukasz Kaiser, and Illia Polosukhin. 2017.
\newblock Attention is all you need.
\newblock \emph{Advances in neural information processing systems}, 30.

\bibitem[{Wang et~al.(2022)Wang, Qin, Zakari, Lu, and Yin}]{wang2022deep}
Hailin Wang, Ke~Qin, Rufai~Yusuf Zakari, Guoming Lu, and Jin Yin. 2022.
\newblock Deep neural network-based relation extraction: an overview.
\newblock \emph{Neural Computing and Applications}, pages 1--21.

\bibitem[{Xiaoyan et~al.(2023)Xiaoyan, Yang, Min, Lingzhi, Rui, Hong, Wai, Ying, and Ruifeng}]{xiaoyan2023comprehensive}
Zhao Xiaoyan, Deng Yang, Yang Min, Wang Lingzhi, Zhang Rui, Cheng Hong, Lam Wai, Shen Ying, and Xu~Ruifeng. 2023.
\newblock A comprehensive survey on deep learning for relation extraction: Recent advances and new frontiers.
\newblock \emph{arXiv preprint arXiv:2306.02051}.

\end{thebibliography}
\end{document}